\documentclass[11pt]{article}

\usepackage[utf8]{inputenc}
\usepackage{graphicx}      % For figures
\usepackage{amsmath,amssymb} % For math symbols
\usepackage{hyperref}      % For links
\usepackage{authblk}       % For author/affiliation block
\usepackage{geometry}      % Adjust page margins
\usepackage{booktabs}
\usepackage{float}
\usepackage{placeins}

\title{Evaluating Large Language Models for Anxiety, Depression, and Stress Detection: Insights into Prompting Strategies and Synthetic Data}
\author[1]{Mihael Arcan}
\author[1]{David-Paul Niland}
\affil[1]{Lua Health, Galway, Ireland}
\date{\today}

\begin{document}

\maketitle

\begin{abstract}
Mental health disorders affect over one-fifth of adults globally, yet detecting such conditions from text remains challenging due to the subtle and varied nature of symptom expression. This study evaluates multiple approaches for mental health detection, comparing Large Language Models (LLMs) such as Llama and GPT with classical machine learning and transformer-based architectures including BERT, XLNet, and Distil-RoBERTa. Using the DAIC-WOZ dataset of clinical interviews, we fine-tuned models for anxiety, depression, and stress classification and applied synthetic data generation to mitigate class imbalance. Results show that Distil-RoBERTa achieved the highest F1 score (0.883) for GAD-2, while XLNet outperformed others on PHQ tasks (F1 up to 0.891). For stress detection, a zero-shot synthetic approach (SD+Zero-Shot-Basic) reached an F1 of 0.884 and ROC AUC of 0.886. Findings demonstrate the effectiveness of transformer-based models and highlight the value of synthetic data in improving recall and generalization. However, careful calibration is required to prevent precision loss. Overall, this work emphasizes the potential of combining advanced language models and data augmentation to enhance automated mental health assessment from text.
\end{abstract}

Mental health challenges impose a substantial burden on individuals and communities globally, with recent data indicating that over 20\% of adults will experience at least one mental disorder during their lifetime \cite{MCGRATH2023668}. Disorders such as depression and anxiety alone account for an estimated \$1 trillion in annual global economic productivity losses \cite{chodavadia2023prevalence}. These challenges are further exacerbated by the subtle and complex ways in which mental health conditions manifest in language, making their detection in text particularly challenging. Symptoms of stress, anxiety, and depression are often conveyed through nuanced and varied expressions, reflecting a wide spectrum of emotional and psychological states that traditional text-based models struggle to interpret and classify accurately. As mental health challenges continue to escalate globally, there is an increasing need for effective methods of early detection and intervention, particularly in everyday contexts such as workplaces \cite{vinson2024towards}. The stigma surrounding mental health remains a significant barrier, often leading to underreporting and delayed care, which further exacerbates the societal impact \cite{Sawma2024.11.22.24317772}. In this context, innovative approaches leveraging technology, such as machine learning (ML) and natural language processing (NLP), hold promise in improving the identification and management of mental health conditions, providing more timely and personalized support to those in need.

Over the past decade, research in NLP and computational social science has explored methods for identifying mental ill-health issues using textual data, including social media content \cite{MONTEJORAEZ2024100654}. Much of this work has focused on building specialized ML models for tasks such as stress detection or depression prediction. While transformer-based models like BERT and XLNet have improved performance, these require fine-tuning for specific tasks, limiting flexibility. Multi-task approaches have also been explored but are often constrained to predefined task sets. Another promising area of research has explored the use of chatbots in mental health services, offering potential for increased accessibility and support. However, many of these tools remain rule-based and lack the sophistication of more advanced AI models, which poses a risk of delivering incorrect interventions that could lead to significant negative outcomes for vulnerable individuals \cite{app14135889}.

ChatGPT\footnote{\url{https://chatgpt.com/}} has transformed how people interact with AI, fueling widespread adoption across industries such as customer service, education, and mental health detection. Built on the GPT AI model family, ChatGPT offers a user-friendly conversational interface that simplifies engagement with Large Language Models (LLMs) for individuals and businesses alike. With ongoing advancements in LLMs like OpenAI's GPT and Meta's Llama models, the potential for AI to address critical challenges, such as mental health detection, is rapidly expanding. LLMs are capable of generating text that closely mirrors real-world data, enabling more accurate identification of mental health conditions by capturing the diverse ways symptoms are expressed \cite{YUAN2024100030,10.1093/bib/bbad493}. These models have the potential to recognize subtle linguistic cues and patterns that traditional diagnostic tools might miss, providing more nuanced insights into individuals' mental health. Additionally, synthetic data generation provides a valuable tool for augmenting datasets, addressing issues such as data scarcity and class imbalance. By introducing artificial examples that replicate the characteristics of real data, synthetic data can enhance model performance and robustness, particularly when access to large, diverse datasets is limited. However, it is critical to ensure that synthetic examples maintain the authenticity and relevance of human expressions to preserve data integrity and avoid potential negative impacts on model performance. Despite these advancements, significant gaps remain in fully understanding the potential of LLMs in the mental health domain, particularly in their ability to comprehend and accurately classify mental health conditions expressed in natural language. Furthermore, the integration of synthetic data into mental health detection systems introduces challenges, including the need to balance diversity and authenticity to ensure effective model training.

This study aims to bridge these gaps by evaluating the performance of LLMs, including OpenAI's GPT-3.5 Turbo and Metas's Llama 3 8B \cite{dubey2024llama}, against classical ML and transformer-based models using the DAIC-WOZ and the Stress Detection dataset. Additionally, we investigate the impact of synthetic data on model performance, exploring how different prompting strategies and training dataset sizes affect classification accuracy. By addressing these questions, we seek to contribute to the development of more reliable and accurate systems for mental health assessment from textual data, ultimately advancing the application of AI in this critical field.

\section{Related Work}

This section reviews advancements in mental health diagnostics and interventions through AI and ML, with a focus on depression, anxiety, stress, workplace toxicity, and large language models (LLMs). Additionally, we explore the role of synthetic data in addressing data scarcity and ethical challenges.

%\subsection{AI for Depression and Anxiety Diagnosis}
To address Major Depressive Disorder (MDD), \cite{fionn_aics18} proposed a passive diagnostic system combining clinical psychology, ML, and conversational systems. Utilizing sequence-to-sequence neural networks, the system enables real-time dialogue with users while monitoring conversations for depression symptoms through specialized classifiers. Despite challenges with small datasets, the study demonstrates the potential for human-like chatbots in real-time mental health support. Additionally, \cite{delahunty_smm4h19} introduced a deep neural network predicting PHQ-4 scores (assessing depression and anxiety) from text. The model leverages the Universal Sentence Encoder and Transformers, incorporating psycholinguistic features to enhance predictions. Though promising, the work highlights challenges related to generalizability and domain-specific applications, especially for social media data. \cite{DBLP:journals/braininf/MilintsevichSD23} focused on detecting MDD using natural language processing (NLP) to create neural classifiers. By analyzing speech transcripts, the approach predicts individual symptoms using symptom network analysis, achieving competitive results in binary diagnosis and depression severity prediction.

%\subsection{Workplace Mental Health and Toxicity Detection}
In workplace settings, \cite{DBLP:conf/emnlp/BhatHABL21} addressed toxic communication through \textit{ToxiScope}, a taxonomy for detecting toxic language patterns in emails. Annotation tasks and ML models revealed the interplay of implicit and explicit toxicity with workplace power dynamics. Future directions emphasize refining detection techniques and investigating biases.

%\subsection{Multimodal Biomarkers for Mental Health}
\cite{Jiang2023.09.11.23295212} explored multimodal biomarkers for psychiatric disorders, utilizing behavioral and physiological signals from remote interviews. Features such as facial expressions, speech, and cardiovascular modulation were analyzed across four tasks, including MDD and self-rated depression detection. Results showed multimodal approaches outperforming unimodal methods, with AUROCs ranging from 0.72 to 0.82, underscoring the value of multimodal biomarkers for scalable mental health assessments.

%\subsection{Applications of Large Language Models (LLMs) in Mental Health}
LLMs, such as GPT-3, GPT-4, and Google's PaLM, have demonstrated transformative potential in mental health. \cite{Stade2023} outlined a roadmap for integrating clinical LLMs in psychotherapy, emphasizing technical, ethical, and practical considerations. The study advocates responsible development and public education on AI's risks and benefits. \cite{chung2023challenges} examined challenges in using LLMs for psychological counseling, such as hallucination, interpretability, and privacy. The authors propose enhancements in fine-tuning and regulatory scrutiny to ensure practical deployment in mental healthcare. Similarly, \cite{galatzerlevy2023capability} evaluated Med-PaLM 2 for predicting psychiatric functioning, achieving near-human performance in depression score prediction across diverse clinical tasks. Social media-based mental health analysis also benefits from LLM advancements. \cite{yang2023mentallama} introduced MentaLLaMA, an interpretable mental health model trained on the \textit{IMHI} dataset, achieving state-of-the-art performance. \cite{xu2023mentalllm} further highlighted the effectiveness of instruction fine-tuning for mental ill-health prediction, positioning LLMs like Mental-FLAN-T5 and Mental-Alpaca as viable alternatives to traditional methods.

Recent research has also explored the use of transformer-based architectures for predicting clinical assessment outcomes from unstructured patient text. For example, \cite{PERISA2024105182} investigated whether transformer models such as BERT and RoBERTa could predict standardized mental health scores—including PHQ-9, GAD-7, PANSS, and the Calgary Depression Scale—using weekly written diaries from patients experiencing a first episode of psychosis. The models achieved high performance on PHQ-9 (F1 = 0.921) and GAD-7 (F1 = 0.945), with mean average errors for PANSS and GAF comparable to clinician interrater variability. These results underscore the capacity of transformer-based models to approximate clinical assessments from natural language, highlighting their potential for early relapse detection and continuous digital monitoring in psychiatric care.

%\subsection{Synthetic Data in Mental Health Research}
Synthetic data addresses data scarcity and privacy challenges in mental health. \cite{10.1016/j.cosrev.2023.100546} demonstrated the utility of SynthNotes in generating realistic, privacy-preserving mental health datasets. Similarly, \cite{6a37997948a6440aaa37fe2be87e1f27} and \cite{doi:10.1177/14604582221077000} developed GAN-based methods for creating statistically accurate synthetic datasets, facilitating robust AI training while maintaining ethical compliance. \cite{lorge2024detectingclinicalfeaturesdifficulttotreat} showcased GPT-3.5 for generating annotated clinical notes to enhance depression diagnostics. Meanwhile, \cite{mori2024algorithmicfidelitymentalhealth} used synthetic data to analyze demographic biases and stress triggers, revealing the potential of AI for equitable research.

%\subsection{Integrating AI with Mental Health Practices}
Data-centric approaches further enhance mental health diagnostics. \cite{10207938} demonstrated the effectiveness of synthetic data in early mental health prediction, using balancing and augmentation strategies to outperform traditional methods. In clinical NLP tasks, models like Distil-RoBERTa and LLMs, like OpenAI's GPT family have excelled in zero-shot tasks, with enhanced prompting strategies improving prediction accuracy \cite{yang2023towards, arcan2024assessmentcomprehendingmentalhealth}. Synthetic data generation remains integral to mental health research, with advancements in frameworks such as MATE-KD \cite{rashid-etal-2021-mate} and the development of task-specific models like SynthNotes contributing to scalable AI solutions.

\section{Methodology}

In this section, we describe the methodology employed for data preprocessing and model training, with a focus on detecting depression and stress across two distinct datasets. Depression detection was conducted using the DAIC-WOZ dataset, while stress detection utilized a combined corpus of Reddit and Twitter data, forming a comprehensive stress detection dataset.

We present the data preprocessing techniques, predictive model training procedures, and prompting strategies designed to enhance the performance of both traditional machine learning models and large language models (LLMs), ensuring each approach is optimized for its respective dataset and task.

\subsection{Depression Classification}
In this section we focus on our approach to leverage the DAIC-WOZ dataset (see Section \ref{subsec:datasets}) to classify depression and anxiety.

\subsubsection{Data Preprocessing}

Our initial dataset is the DAIC-WOZ, which comprises transcribed clinical interviews collected using a Wizard-of-Oz approach for 142 patients. The interviews, centered on general conversation topics, were conducted entirely within the United States. For each patient, the dataset includes a transcript of their interview and corresponding PHQ-8 scores, with bot statements excluded to retain only patient responses. PHQ-8 scores were mapped to PHQ-2 scores, and GAD-2 scores were inferred using the methodology described by \cite{johansson2013depression}.

From these data, we derived PHQ-4 scores for each participant, comprising four variables corresponding to the four items in the PHQ-4 questionnaire. Generalized Anxiety Disorder items 1 and 2 (GAD-1, GAD-2) assess anxiety, while Patient Health Questionnaire items 1 and 2 (PHQ-1, PHQ-2) assess depression. The PHQ-4 scores classify participants into one of four output categories: none, mild, moderate, or severe. Each participant is assigned a single overall PHQ-4 score, broken down into the four categories: GAD-1, GAD-2, PHQ-1, and PHQ-2, regardless of the number of messages in the dataset.

To enhance the contextual information associated with each participant's PHQ-4 score, we concatenated individual messages from the same participant, limiting each observation to a maximum of 50 words. Importantly, only messages from the same individual were combined; no messages from different participants were merged.

After completing the concatenation process and removing duplicates and missing values, the dataset consisted of 28,186 observations in the training set and 8,710 in the test set.

\subsubsection{Transformer-based Models and XGBoost Training}

Based on the preprocessing of the DAIC-WOZ dataset described above, we fine-tuned several transformer-based models, including BERT, DistilBERT, RoBERTa, DistilRoBERTa, and XLNet (see Section \ref{subsec:models}). These models were selected for their proven ability to capture complex linguistic patterns and contextual information, which are essential for natural language understanding tasks such as mental health detection. The fine-tuning process involved initializing each model with pre-trained weights and adapting them to the specific task of predicting psychological distress through optimization on the DAIC-WOZ dataset.

The DAIC-WOZ dataset, which comprises clinical interviews aimed at assessing psychological distress, provided a robust and contextually rich training corpus. Each input text, corresponding to an individual's responses, was mapped to their PHQ-4 score, enabling supervised learning. Weighted F1 score was used as the primary metric to evaluate model performance.

For the XGBoost model, the DAIC-WOZ dataset was first converted into dense vector representations using a pre-trained Transformer model. The model training involved hyperparameter optimization through a grid search within a five-fold cross-validation framework. Key parameters, such as the number of trees, maximum depth, and learning rate, were tuned to maximize performance. The best-performing XGBoost model was selected based on the weighted F1 score, ensuring robustness across imbalanced classes. To ensure reproducibility, all models were trained on the same dataset splits, and random seeds were set to control variability in training outcomes. 

\subsubsection{Prediction Using LLM Prompting}

To evaluate and compare the predictive capabilities of LLMs, like GPT and Llama, against transformer-based models and XGBoost, we explored various prompting strategies designed to elicit specific responses. We prompted Llama 3 8B (see Section \ref{subsubsec:llama3}) with five distinct queries and assessed their performance based on the generated outputs. Variations in the prompts included differences in lexicalized outputs, definitions, and prompt lengths. These strategies were specifically tailored to address the PHQ or GAD questions, resulting in a total of 20 unique prompts. By comparing the performance of these strategies with the predictions made by transformer models and XGBoost, we aimed to evaluate the relative strengths and weaknesses of each approach in predicting mental health outcomes.

The differences between prompt versions were minor, such as variations in phrasing like "on a scale of zero to three" versus "on a scale of 0 to 3," or in prompt length, with some being brief (e.g., Version 3 in Table \ref{tab:prompting_examples1}) and others more detailed (e.g., Version 5 in Table \ref{tab:prompting_examplesV5}), including examples of messages evaluated based on specific GAD or PHQ questions. The goal was for the LLM to return a number corresponding to the relevant PHQ or GAD question, and regular expressions were employed to handle the outputs. This involved extracting the first digit from the string or converting numerically expressed words (e.g., "zero") into integers. For instance, if the chatbot response was \textit{"This message looks like it is showing zero levels of anxiety"} our system would return "0," facilitating the analysis and evaluation process.

\begin{table}[]
    \small
    \begin{tabular}{l|p{154mm}}
    \toprule
        Version 1 \\
        \midrule
        GAD-1 & On a scale of \textbf{zero} to \textbf{three}, estimate how anxious this person is with zero being a little anxious and three being incredibly anxious:+\texttt{TEXT}\\
        GAD-2 & On a scale of \textbf{zero} to \textbf{three}, estimate how worried this person is with zero being a little worried and three being incredibly worried:+\texttt{TEXT}\\
        PHQ-1 &  On a scale of \textbf{zero} to \textbf{three}, estimate how depressed this person is with zero being a little depressed and three being incredibly depressed:+\texttt{TEXT}\\
        PHQ-2 & On a scale of \textbf{zero} to \textbf{three}, estimate if this person is expressing no interest in previously enjoyed activities with zero being a little and three being incredibly:+\texttt{TEXT}\\
        \midrule

%        Version 2 \\p
%        \midrule
%        GAD-1 & On a scale of zero to three, estimate how anxious this person is with zero being a little anxious and three being incredibly anxious \\
%        GAD-2 & On a scale of zero to three, estimate how worried this person is with zero being a little worried and three being incredibly worried \\
%        PHQ-1 & On a scale of zero to three, estimate how depressed this person is with zero being a little depressed and three being incredibly depressed \\
%        PHQ-2 & On a scale of zero to three, estimate if this person is expressing no interest in previously enjoyed activities with zero being a little and three being incredibly \\
%        \midrule
%        \midrule
        Version 2 \\
        \midrule
        GAD-1 & On a scale of \textbf{0} to \textbf{3}, estimate how anxious this person is with 0 being a little anxious and 3 being incredibly anxious:+\texttt{TEXT}\\
        GAD-2 & On a scale of \textbf{0} to \textbf{3}, estimate how worried this person is with 0 being a little worried and 3 being incredibly worried:+\texttt{TEXT}\\
        PHQ-1 & On a scale of \textbf{0} to \textbf{3}, estimate how depressed this person is with 0 being a little depressed and 3 being incredibly depressed:+\texttt{TEXT}\\
        PHQ-2 & On a scale of \textbf{0} to \textbf{3}, estimate if this person is expressing no interest in previously enjoyed activities with 0 being a little and 3 being incredibly:+\texttt{TEXT}\\
        \midrule

        Version 3 \\
        \midrule
        GAD-1 & On a scale of \textbf{zero} to \textbf{three}, rate the anxiety of this message:+\texttt{TEXT}\\
        GAD-2 & On a scale of \textbf{zero} to \textbf{three}, rate the worry in this message:+\texttt{TEXT}\\
        PHQ-1 & On a scale of \textbf{zero} to \textbf{three}, rate the depression in this message:+\texttt{TEXT}\\
        PHQ-2 & On a scale of \textbf{zero} to \textbf{three}, rate the interest in previously enjoyed activities in this message:+\texttt{TEXT}\\
        \midrule
        Version 4 \\
        \midrule
        GAD-1 & \textbf{Generalised anxiety disorder is a mental health illness that is defined by people having feelings of excessive anxiety.} On a scale of zero to three, rate the anxiety in this message:+\texttt{TEXT}\\
        GAD-2 & \textbf{Generalised anxiety disorder is a mental health illness that is defined by people having feelings of excessive worry.} On a scale of zero to three, rate the anxiety in this message:+\texttt{TEXT}\\
        PHQ-1 & \textbf{Depression, or major depressive disorder is a mental health illness that is categorised by people feeling down, depressed or hopeless.} On a scale of zero to three, rate the depression in this message:+\texttt{TEXT}\\
        PHQ-2 & \textbf{Depression, or major depressive disorder is a mental health illness that is categorised by people having little interest or pleasure in doing things.} On a scale of zero to three, rate the depression in this message:+\texttt{TEXT}\\
    \bottomrule
    \end{tabular}
    \caption{Examples for different prompting versions for GAD and PHQ questions.}
    %\vspace{100mm}
    \label{tab:prompting_examples1}
\end{table}

\begin{table}[]
    \small
    \centering
    \begin{tabular}{c|p{154mm}}
    \toprule
        %\begin{tiny}Version 5\end{tiny} \\
        \midrule
        GAD-1 & \small Generalised anxiety disorder is a mental health illness that is defined by people having feelings of exessive anxiety. \textbf{The anxiety in this example is rated zero:} "like uh a a a guy that likes to see different sights and go different places uh like different i'm a big movie person so i like going to theatres stuff like that try to free myself from the situation to clear my mind usually take a walk or something" \textbf{The anxiety in this example is rated one:} "got stuck out here then i had my baby and i just stayed be doing a little bit of everything a good night's sleep i grew up not getting to know who my mom was out there just on the food basis they like to give you a lotta food" \textbf{The anxiety in this example is rated two:} "it was it was great and now it's just okay i'm hardworking i'm uh i care about everyone i yep i suppose going on a trip to canada with my daughter was very memorable i love la i love the traffic on the four o five i love the beach" %The anxiety in this example is rated three: "regret right now is um that's just how she is value we were always what what's the word i'm looking for um also the weather was just it was absolutely awful there and it was just it was very difficulties in my life she's very touchy i would say in" \textbf{The anxiety in this example is rated zero:} "like uh a a a guy that likes to see different sights and go different places uh like different i'm a big movie person so i like going to theatres stuff like that try to free myself from the situation to clear my mind usually take a walk or something" \textbf{The anxiety in this example is rated one:} "got stuck out here then i had my baby and i just stayed be doing a little bit of everything a good night's sleep i grew up not getting to know who my mom was out there just on the food basis they like to give you a lotta food". \textbf{The anxiety in this example is rated two:} "mean by them um i garden a little bit i read a little bit i watch tv i'm a basketball fan i go to the beach when the weather is nice go to the pool oh i don't know a year or so we're high energy people really great and. 
        \textbf{The anxiety in this example is rated three:} "regret right now is um that's just how she is value we were always what what's the word i'm looking for um also the weather was just it was absolutely awful there and it was just it was very difficulties in my life she's very touchy i would say in." On a scale of zero to three, rate the anxiety in this message:+\texttt{TEXT} \\ 
        GAD-2 & \small Generalised anxiety disorder is a mental health illness that is defined by people having feelings of excessive worry. \textbf{The worry in this example is rated zero:} "i've added that to my life she's a great woman and uh that started a future in in athletics and that's what i wanted to do so i accomplished my goals so at the time it was a great a great thing and it still is a great thing it's" \textbf{The worry in this example is rated one:} "lot of stress every day just everyday stress uh yeah pretty much that's what triggers it play like games like i like to play video games those are fun stuff like that okay thanks eh my family it's just eh we're not like too too close we kinda argue a" \textbf{The worry in this example is rated three:} "guess that and not and then there's you i don't completely lose it for days at a time when i can get really into that and sort of shut out the rest of the world um the music and the the thoughts just kind of it really makes me a" On a scale of zero to three, rate the anxiety in this message:+\texttt{TEXT}\\
        PHQ-1 & \small Depression, or major depressive disorder is a mental health illness that is categorised by people feeling down, depressed or hopeless. \textbf{The depression in this example is rated zero:} "like uh a a a guy that likes to see different sights and go different places uh like different i'm a big movie person so i like going to theatres stuff like that try to free myself from the situation to clear my mind usually take a walk or something" \textbf{The depression in this example is rated one:} "of my friend i wish i would've handled his sister a little differently as far as the dirt it has been yes it was uh it's very close um no i have not we've always maintained our friendship um i can't recall one off hand no problem and that's all" \textbf{The depression in this example is rated two:} "fight or anything or yeah any of that my grandma she's always giving me encouragement and um she's a therapist a licensed therapist so she's always you know made it really really um yeah i'm okay yeah um i just try to stay positive i try to think like okay" \textbf{The depression in this example is rated three:} "to my son my son and daughter-in-law and my daughter and i went out to have hawaiian food got together it's always fun to get together we got together at my house on christmas i enjoyed just getting together with them i guess the newness wore off that just have" On a scale of zthat’sero to three, rate the depression in this message:+\texttt{TEXT}\\
        PHQ-2 & \small Depression, or major depressive disorder is a mental health illness that is categorised by people having little interest or pleasure in doing things. \textbf{The depression in this example is rated zero:} "like uh a a a guy that likes to see different sights and go different places uh like different i'm a big movie person so i like going to theatres stuff like that try to free myself from the situation to clear my mind usually take a walk or something" \textbf{The depression in this example is rated one:} "of my friend i wish i would've handled his sister a little differently as far as the dirt it has been yes it was uh it's very close um no i have not we've always maintained our friendship um i can't recall one off hand no problem and that's all" \textbf{The depression in this example is rated two:} "um feel uninhibited uninhibited and i was better prepared about three years ago um i was happy that he was safe try not to she has a house i mean a roof over her head to resort to the situation that he was in um because and to have been" \textbf{ The depression in this example is rated three:} "i don't sleep well um well i start to like cry a lot and i start to get really irritable um i argued with my mom and sister yesterday it was just something stupid over like yeah i was like okay well this is the problem and it i just" $<</SYS>>$ On a scale of zero to three, rate the depression in this message:+\texttt{TEXT}\\
    \bottomrule
    \end{tabular}
    \caption{Examples for version 5 prompting for the GAD and PHQ questions.}
    \label{tab:prompting_examplesV5}
\end{table}

%\subsubsection{ChatGPT}
%For ChatGPT, we used the same prompting structure that worked best for Llama-2, which was Version 3 (see Table~\ref{tab:prompting_examples1}). All other aspects of evaluation and extracting information from the chatbot responses were the same for ChatGPT as it was for Llama-2. 

Furthermore, we utilized the prompting structure that proved most effective for Llama 3 8B, known as Version 3 (Table~\ref{tab:prompting_examples1}), for GPT-3.5 Turbo evaluation. This allowed us to assess the relative performance of both models under consistent conditions. All other evaluation aspects, including the methods for information extraction from chatbot responses, remained consistent across both GPT and Llama.

\subsubsection{LLM Fine-Tuning}

In addition to comparing the default GPT and Llama 3 LLMs, we conducted a separate experiment to assess the impact of fine-tuning GPT-3.5 Turbo. This experiment comprised two sub-tasks. The first sub-task involved a multiclass classification task based on the PHQ-4 questionnaire, with four output classes: 0 for none, 1 for mild, 2 for moderate, and 3 for severe. The second sub-task involved a binary classification of PHQ-4, where non-symptomatic and mildly symptomatic instances were grouped into Class 0, and moderate to severe instances were assigned to Class 1 (see Table~\ref{tab:prompting_examples_experiment2}). We fine-tuned GPT 3.5 turbo on the DAIC-WOZ dataset, whereby the training set consisted of 28,186 observations and the test set contained 8,710. 

%In the multiclass experiment, we aimed to replicate the structure of the Version 3 prompts from previous experiments as closely as possible. However, when we directly adapted the Version 3 prompts by only changing the target to a value between 0 and 1, the model produced outputs within this range, which was not the goal(e.g., 0.8). To better align the prompts with the specific requirements of the task, we made further adjustments.

\begin{table}[]
    \begin{tabular}{l|p{130mm}}
    \toprule
        Binary PHQ-4 & Rate the anxiety and depression of this message with a score of 0 for no anxiety or depression or 1 for anxiety or depression. Only output a 0 or a 1: +\texttt{TEXT}\\
        \midrule
        Multiclass PHQ-4 & On a scale of 0 to 3, rate the severity of anxiety and depression in this message: +\texttt{TEXT}\\
    \bottomrule
    \end{tabular}
    \caption{Examples of the adapted prompting versions for GPT 3.5 Turbo fine-tuning.} %\vspace{20mm}
    \label{tab:prompting_examples_experiment2}
\end{table}

\subsection{Synthetic Data Generation and Training}

In contrast to training transformer-based models and XGBoost models on the existing DAIC-WOZ dataset, we investigate the use of generative LLMs to create synthetic datasets aimed at enhancing classification models for depression and stress detection. This approach addresses common challenges in mental health datasets, such as data scarcity and class imbalance \cite{salmi2024handling}. Tables \ref{tab:synt_prompt_depression} and \ref{tab:synt_prompt_stress} present the prompt configurations and example setups used for synthetic data generation.

\begin{table*}[t]
    \parbox{.45\linewidth}{
    \centering
    \setlength{\tabcolsep}{12pt}
    \begin{tabular}{c|cccc} 
    \toprule
    Prompt & Shots &\# Prompts \\
    \midrule
    Zero-Shot-Basic & 0 & 5  \\ 
    Few-Shot-Basic & 5 &5\\
    Few-Shot-Extended & 5 & 15 \\
    Few-Shot-Enhanced & 7 & 5 \\ %%FewMoreBasic
    \bottomrule
    \end{tabular}
    \caption{Prompting structure for depression.}
    \label{tab:synt_prompt_depression}
    \vspace{2.08mm}
    }
    \hfill
    \parbox{.46\linewidth}{
    \begin{tabular}{c|ccc} 
    \toprule
    Prompt & Shots &\# Prompts & Topics\\
    \midrule
    Zero-Shot-Basic & 0 & 5& -\\ 
    Few-Shot-Basic & 5 &5&- \\
    \midrule
    General & 0 & 1&5\\
    Specific & 0 & 1&43\\
    \bottomrule
    \end{tabular}
    \caption{Prompting structure for stress.}
     \label{tab:synt_prompt_stress}
    }
    \label{tab:performance_metrics}
\end{table*}

\subsubsection{Data Preprocessing}

Similarly, we utilize the DAIC-WOZ dataset for synthetic data generation, incorporating individuals' conversations in few-shot scenarios when prompting GPT-3.5 with real-world examples. This approach aims to enhance the model's ability to generalize to diverse conversational patterns.

In our focus on stress detection, we leverage the Stress Detection dataset (see Section \ref{subsubsec:sd_dataset}), which includes both text and metadata from Reddit posts, in combination with the Twitter Full dataset, to create a comprehensive stress detection corpus. The combined dataset was then partitioned into 9,249 training examples and 2,313 test examples, providing a robust foundation for evaluation.

\begin{table}[]
    \centering
    \small
    \begin{tabular}{p{160mm}}
    \toprule
         \textbf{Zero-Shot-Basic (for Depression)} \vspace{-1mm}
    %\midrule
        \begin{enumerate}
            \item  Depression is a major problem affecting many people around the world. Generate some examples of messages from someone who is likely to be suffering from depression: \vspace{-2mm}
            \item Depression is a common and often unnoticed disorder. Generate example messages from someone who could be suffering from depression: \vspace{-2mm}
            \item Generate messages from the first person from someone who is experiencing depression:  \vspace{-2mm}
            \item Generate messages that suggest that the person who wrote them might be suffering from major depressive disorder: \vspace{-2mm}
            \item Depression is often categorized as feelings of sadness, hopeless and without interest in previously enjoyed activites. Generate examples from the first person of someone who might be experiencing depression: \vspace{1mm}
        \end{enumerate} 
    %\midrule
        \\ \textbf{Few-Shot-Basic (for  Depression)} \vspace{-1mm}
    %\midrule
        \begin{enumerate}
            \item  Here are some examples of messages from someone who is suffering from depression: \texttt{5 Examples} \vspace{-2mm}
            \item  Here are some examples messages of someone who is likely to be suffering from major depressive disorder: \texttt{5 Examples} ... \vspace{-2mm}
            \item  This person is probably suffering from depression: \texttt{5 Examples} ... \vspace{-2mm}
            \item  These are messages from someone who is likely showing moderate or severe levels of depression: \texttt{5 Examples} ... \vspace{-2mm}
            \item  Major depressive disorder (MDD), also known as clinical depression, is a mental disorder characterized by at least two weeks of pervasive low mood, low self-esteem, and loss of interest or pleasure in normally enjoyable activities. Here are some messages from someone who is likely suffering from derpession: \texttt{5 Examples} ... \vspace{1mm}
         \end{enumerate}
    %\midrule
        \\ \textbf{Few-Shot-Extended (Depression)} \vspace{-1mm}
    %\midrule
         \begin{enumerate}
            \item  Depression is a major problem affecting many people around the world. Here are some examples of messages from people who are likely suffering from depression: \texttt{5 Examples} ... \vspace{-2mm}
            \item  Depression is a common and often unnoticed disorder. Here are some examples of people who could be suffering from depression: \texttt{5 Examples} ... \vspace{-2mm}
            \item  These messages are from someone who is experiencing depression: \texttt{5 Examples} ... \vspace{-2mm}
            \item  These messages suggest that the person might be suffering from major depressive disorder: \vspace{-2mm}
            \item  These messages are from a person who is experiencing low mood: \texttt{5 Examples} ... \vspace{-2mm}
            \item  In these messages, there are indications that the individual is likely experiencing symptoms of major depressive disorder: \texttt{5 Examples} ...
            \item  These messages reflect someone's experience of feeling overwhelmed by sadness and hopelessness: \texttt{5 Examples} ...\vspace{-2mm}
            \item  The author of these messages likely feels sad, hopeless and without interest in previously enjoyed activites: \texttt{5 Examples} ... \vspace{-2mm}
            \item  These texts suggest that the person is experiencing a loss of interest in activities they used to enjoy, a common symptom of depression: \texttt{5 Examples} ... \vspace{-2mm}
            \item  These messages indicate persistent feelings of worthlessness or guilt, often associated with depression: \texttt{5 Examples} ... \vspace{-2mm}
            \item  These texts suggest that the person is experiencing changes in appetite or weight, which can be symptoms of depression: \texttt{5 Examples} ... \vspace{-2mm}
            \item  These messages indicate that the individual is experiencing significant changes in sleep patterns, which are often linked to depression: \texttt{5 Examples} ... \vspace{-2mm}
            \item  Persistently expressing thoughts of low mood or lack of interest in doing previously enjoyed activitites, indicate moderate or severe depression. The person who wrote these messages could be suffering from depression: \texttt{5 Examples} ... \vspace{-2mm}
            \item  The following messages are examples of someone who might be having difficulties in concentration, decision-making, or memory, common cognitive symptoms of depression: \texttt{5 Examples} ... \vspace{-2mm}
            \item  These texts suggest that the person is withdrawing from social interactions, a common symptom of depression: \texttt{5 Examples} ... 
        \end{enumerate}
    \\ \bottomrule 
    \end{tabular}
    \vspace{-4mm} 
    \caption{Examples used for synthetic data generation for depression.}
    \label{tab:synthetic_depression}
\end{table}

\subsubsection{Prompting Strategies for Depression}

To generate data tailored for both depression and stress contexts, we employed a range of prompting strategies. These strategies illustrated in Table~\ref{tab:synthetic_depression} aimed to capture diverse aspects of mental health through varied approaches. To enhance data diversity without using examples from the base dataset, we applied a \textbf{Zero-Shot-Basic Prompt} strategy. This involved five prompts designed to generate data for depression and non-depression scenarios without any specific examples from the dataset, aiming to capture a wider range of expressions. The \textbf{Few-Shot-Basic Prompts} strategy involved designing five distinct prompts, each with five examples, to illustrate scenarios of depression and non-depression. The prompts were carefully crafted to guide the generation of representative messages for each context. Building on the Few-Shot-Basic approach, the \textbf{Few-Shot-Extended Prompts} introduced ten additional prompts. This expansion aimed to increase the diversity of generated data by providing a broader range of examples, further refining the context of both depression and non-depression. The \textbf{Few-Shot-Enhanced Prompts} were an extension of the Few-Shot-Basic strategy, incorporating a larger number of examples from the DAIC-WOZ dataset. This approach aimed to refine the generated messages by offering more detailed guidance to the language model.

\begin{table}[]
    \centering
    \small
    \begin{tabular}{p{160mm}}
    \toprule
        \textbf{Zero-Shot-Basic (Stress)}\\
    \midrule
        Stress is a major problem affecting many people around the world. Generate some examples of messages from someone who is likely to be suffering from stress: \\
        Stress is a common and often unnoticed disorder. Generate example messages from someone who could be suffering from stress: \\
        Generate messages from the first person from someone who is experiencing stress: \\
        Generate messages that suggest that the person who wrote them might be suffering from stress: \\
    \midrule
        \textbf{Few-Shot-Basic (Stress)}\\
    \midrule
        Here are some examples of messages from someone who is suffering from stress: \texttt{5 Examples} ...\\
        Here are some examples messages of someone who is likely to be suffering from stress: \texttt{5 Examples} ...\\
        This person is probably suffering from stress: \texttt{5 Examples} ...\\
        These are messages from someone who is likely showing moderate or severe levels of stress: \texttt{5 Examples} ...\\
    \midrule 
        \textbf{General Topic Prompts}\\
    \midrule
        These are the criteria of different stress risk level: Risk Level=No Stress: I do not see evidence that this person is suffering from stress. Risk Level=Stress: I believe this person is at high risk of suffering from stress. \\
        
        Your task is to generate a text for each of the following "topics" with different Risk levels.
        	\textbf{1-Crises/catastrophes,
        	2-Major life events,
        	3-Daily hassles/microstressors,
        	4-Ambient stressors,
        	5-Organisational stressors,}
        
        Provide the answers with the following columns: text, topic, risk level. Risk level criteria: "{Stress}"\\
    \midrule
        \textbf{Specific Topic Prompts}\\
    \midrule 
        These are the criteria of different stress risk level: Risk Level=No Stress: I do not see evidence that this person is suffering from stress. Risk Level=Stress: I believe this person is at high risk of suffering from stress. Your task is to generate a text for each of the following "topics" with different Risk levels.
        	\textbf{1-Death of a spouse,
        	2-Divorce,
        	3-Marital separation,
        	4-Imprisonment,
        	5-Death of a close family member,
        	6-Personal injury or illness,
        	7-Marriage,
        	8-Dismissal from work,
        	9-Marital reconciliation,
        	10-Retirement,
        	11-Change in health of family member,
        	12-Pregnancy,
        	13-Sexual difficulties,
        	14-Gain a new family member,
        	15-Business readjustment,
        	16-Change in financial state,
        	17-Death of a close friend,
        	18-Change to different line of work,
        	19-Change in frequency of arguments,
        	20-Major mortgage,
        	21-Foreclosure of mortgage or loan,
        	22-Change in responsibilities at work,
        	23-Child leaving home,
        	24-Trouble with in-laws,
        	25-Outstanding personal achievement,
        	26-Spouse starts or stops work,
        	27-Begin or end school,
        	28-Change in living conditions,
        	29-Revision of personal habits,
        	30-Trouble with boss,
        	31-Change in working hours or conditions,
        	32-Change in residence,
        	33-Change in schools,
        	34-Change in recreation,
        	35-Change in church activities,
        	36-Change in social activities,
        	37-Minor mortgage or loan,
        	38-Change in sleeping habits,
        	39-Change in number of family reunions,
        	40-Change in eating habits,
        	41-Vacation,
        	42-Minor violation of law}
        
        Provide the answers with the following columns: text, topic, risk level. Risk level criteria: "{Stress}"\\
    \bottomrule
    \end{tabular}
    \caption{Examples used for synthetic data generation for stress.}
    \label{tab:synthetic_stress}
\end{table}

\subsubsection{Prompting Strategies for  Stress}

%The \textbf{General Prompts} addressed a range of stress-related topics, including major life events, daily hassles, and organizational stressors. We included topics related to stress to provide insights into the psychological states associated with various stressors. These topics encompassed major life events, crises/catastrophes, daily hassles/microstressors, ambient stressors, and organizational stressors. This approach was designed to capture various psychological states associated with different stressors.

Similar to the synthetic data generation for depression, we leverage a \texttt{Zero-Shot} and \texttt{Few-Shot} prompting strategy (Table \ref{tab:synthetic_stress}). In addition to these strategies outlined above, we employed the \textbf{General Prompt} strategy, which incorporated a broad range of stress-related topics for generating synthetic data aimed at stress detection. This strategy encompassed major life events, daily hassles, and organizational stressors, integrating these topics into the prompts for the LLMs to provide insights into the psychological states linked to various stressors. Specifically, the prompts addressed crises, microstressors, ambient stressors, and organizational stressors, offering a comprehensive perspective on the psychological impact of different types of stress.

Similar to the general prompts, the \textbf{Specific Prompts} were based on the \textit{Holmes and Rahe Stress Scale} \cite{HOLMES1967213} and focused on 43 stress-inducing life events. These included significant events such as the death of a spouse and divorce, offering a targeted approach to generating stress-related data.\footnote{Full list of Holmes and Rahe Stress Scale:\\ \url{https://www.stress.org/wp-content/uploads/2024/02/Holmes-Rahe-Stress-inventory.pdf}} As with the general prompts, these topics were incorporated into the LLM prompts to ensure focused and relevant data generation.

\subsubsection{Training}

We first trained the depression and stress models using only the baseline datasets, i.e., the DAIC-WOZ and Stress Detection datasets, without any synthetic data, in order to establish baseline performance. Synthetic data were then generated to augment these baseline datasets.

For depression, we created four synthetic data versions with 10,000 generated examples each, two versions with 100,000 examples, and two versions with 500,000 examples. From the 10,000-example datasets, the top two prompting strategies were selected based on F1 performance and extended to the 100,000 and 500,000 datasets. Training cycles were conducted to combine the four 10,000-example datasets into a single 40,000-example dataset, in order to assess the combined value of all prompting strategies. Similarly, the two 100,000-example datasets were merged into a 200,000-example synthetic dataset, and the two 500,000-example datasets were consolidated into 1 million synthetic data points.

For stress, we used two training sets generated from the original prompting strategies, i.e., \texttt{Zero-Shot} and \texttt{Few-Shot} strategies, each containing 100,000 examples, as well as two additional sets generated from general and specific strategies, also with 100,000 examples each. Consistent with the depression approach, these datasets were merged to create 200,000 examples for each strategy grouping. All generated data were integrated with the Stress Detection training set.

\section{Experimental Setup}

This section provides insights at the datasets and models utilized in our work. Additionally, it details the evaluation metrics employed to present the outcomes effectively.

\subsection{Datasets}
\label{subsec:datasets}

\subsubsection{DAIC-WOZ}
\label{subsub:daicwoz}
Within this comparison, we leveraged the DAIC-WOZ dataset \cite{gratch-etal-2014-distress}, which comprises clinical interviews aimed at aiding the assessment of psychological distress conditions like anxiety, depression, and post-traumatic stress disorder. These interviews were gathered as part of a broader initiative to develop a computer-based system that conducts interviews with individuals and recognises verbal and nonverbal cues associated with mental health issues. Specifically, it encompasses data from Wizard-of-Oz interviews, where an animated virtual interviewer named Ellie, under the control of a human interviewer in a separate location, conducted the interviews. The data, which consists of 189 interaction sessions, each lasting between 7 to 33 minutes, has been originally transcribed and annotated to encompass a range of verbal and non-verbal characteristics.

%Previous research has determined that a score of 3 or higher on the Depression subscale is a reasonable threshold for identifying potential cases of depression. A score at or above 3 indicates a positive result, warranting further assessment with the PHQ-9 or a referral to mental health services. Similarly, a score of 3 or greater on the Anxiety subscale serves as a sensible cutoff point, signaling the need for further evaluation using the GAD-7 or a mental health referral.
%Elevated scores can indicate the possibility of various disorders, including but not limited to Bipolar I, Bipolar II, Cyclothymia, Dysthymia, Generalized Anxiety Disorder, Social Anxiety Disorder, Panic Disorder, Obsessive-Compulsive Disorder, or Personality Disorders. It's important for patients to understand that a negative screening result does not rule out the presence of a disease; rather, it suggests that the likelihood of having the disease is low.

\subsubsection{Stress Detection dataset}
\label{subsubsec:sd_dataset}
The Stress Detection dataset\footnote{\url{https://www.kaggle.com/datasets/tihsrahly/stress-detection-dataset}} integrates Reddit and Twitter data to analyze stress-related content. The data, labeled as stress-positive or stress-negative, was collected using the Reddit API and Twitter API V2 from September 2019 to September 2021, capturing trends during the COVID-19 pandemic. Reddit data was collected from subreddits like \textit{r/Stressed} and \textit{r/Depression} for stress-positive examples, and \textit{r/Happy} and \textit{r/Wholesome} for stress-negative examples. Twitter data was gathered using hashtags such as \textit{\#Stress} and \textit{\#FeelingStressed} for stress-positive examples, and \textit{\#Happiness} and \textit{\#Joy} for stress-negative examples. Reddit data was detailed and structured, while Twitter data was shorter and less grammatically structured.

%\subsubsection{PHQ-4 Questionnaire}
%The Patient Health Questionnaire-4 (PHQ-4) \cite{KroenkePhq4} was developed to address the challenge posed by the high prevalence of anxiety and depression in the general population. Since these two mood disorders often co-occur and individuals with these conditions may struggle with fatigue or difficulty concentrating, the PHQ-4 offers a concise and accurate assessment tool. The PHQ-4 consists of four questions, each answered on a four-point Likert-type scale. It serves the purpose of providing a very brief yet precise measurement of the fundamental symptoms associated with depression and anxiety. It combines a two-item measure for depression (PHQ–2), which focuses on core depressive criteria, and a two-item measure for anxiety (GAD–2), both of which have been independently proven to be effective screening tools. The total PHQ–4 score complements the scores of these subscales, offering an overall assessment of symptom burden, functional impairment, and disability. While an elevated PHQ–4 score is not diagnostic, it serves as an indicator for further evaluation to confirm the presence or absence of a clinical disorder that requires treatment.

\subsection{Models}
\label{subsec:models}

\subsubsection{Transformer Models}
For a comparison to LLMs, we leverage the Transformer models \cite{NIPS2017_3f5ee243}, which rely on a self-attention mechanism, allowing it to capture contextual dependencies in input sequences efficiently. The model consists of an encoder-decoder structure, with multi-head self-attention layers enabling parallelised processing of input tokens. Positional encoding is used to provide information about the token's position in the sequence. The Transformer's attention mechanism facilitates capturing long-range dependencies, making it highly effective for tasks requiring context understanding. Within this work, we compare the BERT and Roberta Transformer models and their distilled versions, i.e. DistilBert and Distil-Roberta. In addition to that we leverage the XLNet models as well. The BERT model \cite{devlin2019bert} is pre-trained on large corpora and can then be fine-tuned for specific natural language processing (NLP) tasks, such as text classification, named entity recognition, and question answering, among others. BERT embeddings have been widely adopted and have significantly improved the state-of-the-art performance in various NLP applications. DistilBERT \cite{sanh2020distilbert} is a distilled version of BERT, offering a more compact and faster alternative for tasks in natural language processing (NLP). Despite having fewer parameters, DistilBERT embeddings can be utilised in various NLP applications, providing a balance between computational efficiency and model performance. RoBERTa \cite{liu2019roberta}, or Robustly optimised BERT approach, is a variant of the BERT (Bidirectional Encoder Representations from Transformers) model, uses dynamic masking during pre-training, removing the Next Sentence Prediction (NSP) objective, and training with larger mini-batches and learning rates. These modifications result in a more robust and efficient model. XLNet \cite{10.5555/3454287.3454804} is a Transformer-based language model, which combines ideas from autoregressive language modeling (as seen in models like GPT) and autoencoding (as in BERT) to capture bidirectional context and maintain long-term dependencies in sequences. Instead of predicting the next word in a sentence, XLNet is trained to predict a permutation of the words. This approach allows the model to consider bidirectional context while preventing it from seeing the entire context during training, enhancing its ability to capture dependencies. For all models, we leverage the base version of the transformer models.

\subsubsection{XGBoost}
\label{subsub:xgboost}

XGBoost (Extreme Gradient Boosting) \cite{Chen:2016:XST:2939672.2939785} is an ML algorithm used for both classification and regression tasks. XGBoost is based on the gradient boosting framework, which is an ensemble learning technique. It builds an ensemble of decision trees sequentially, where each tree corrects the errors made by the previous ones. The model uses a customizable objective function that needs to be optimised during training. For regression tasks, the objective is often mean squared error (MSE), while for classification tasks, it can be log loss (binary or multiclass). To prevent overfitting, XGBoost incorporates L1 (Lasso) and L2 (Ridge) regularization techniques into the objective function.

\subsubsection{OpenAI GPT}
%GPT-3 (Generative Pre-trained Transformer) follows the decoder-only Transformer architecture and employs attention mechanisms, allowing it to focus on the most relevant segments of input text, using an extensive context of 2048 tokens and an unprecedented 175 billion parameters. The model exhibited remarkable zero-shot and few-shot learning capabilities across various tasks. The training data for GPT-3 is primarily sourced from a filtered version of Common Crawl, contributing to 60\% of the weighted pre-training dataset, comprising 410 billion byte-pair-encoded tokens. Other data sources include 19 billion tokens from WebText2, 12 billion tokens from Books1, 55 billion tokens from Books2, and 3 billion tokens from Wikipedia. GPT-3 was trained on a vast corpus of text and has demonstrated proficiency in programming languages such as CSS, JSX, and Python, among others. Open AI, the commercial company behind ChatGPT, allows for fine-tuning their LLM models to a specific domain, which allows for adaptation to particular a use case. Furthermore, the fine-tuning step allows for greater standardisation and control of outputs as well.

Within this work, we leveraged the GPT-3.5 Turbo model, a large language model based on the decoder-only Transformer architecture, employing attention mechanisms to focus on the most relevant segments of input text. It supports a context length of up to 2048 tokens and features 175 billion parameters, enabling exceptional zero-shot and few-shot learning capabilities across diverse tasks. The model was trained on a vast corpus of text, primarily sourced from a filtered version of Common Crawl (60\% of the pre-training dataset), alongside datasets such as WebText2, Books1, Books2, and Wikipedia, comprising a total of 410 billion byte-pair-encoded tokens. Additionally, GPT-3.5 demonstrates proficiency in programming languages, including CSS, JSX, and Python. The model's adaptability is enhanced through fine-tuning, allowing customization for specific domains and ensuring greater standardization and control over outputs. GPT-3.5 has been employed for two key tasks: classification and synthetic data generation.

\subsubsection{LLama 3}
\label{subsubsec:llama3}
%LLaMA-1 (Large Language Model Meta AI) \cite{touvron2023llama1} is a series of large language models developed by Meta AI. The initial release included models with varying parameter sizes: 7 billion, 13 billion, 33 billion, and 65 billion parameters. LLaMA employs the Transformer architecture with some architectural differences, including the use of SwiGLU activation functions, rotary positional embeddings, and root-mean-squared layer normalization. The foundational models were trained on a vast dataset comprising 1.4 trillion tokens from various publicly available sources. Human annotators were involved in AI alignment by providing prompts and evaluating model outputs, and reinforcement learning from human feedback (RLHF) was employed with a new technique based on Rejection sampling followed by Proximal Policy Optimization (PPO). Additionally, LLaMA aimed to improve multi-turn consistency in dialogues using the "Ghost attention" technique during training to respect system messages throughout conversations. LLaMA-2 \cite{touvron2023llama2} remains mostly consistent with that of LLaMA-1 models, with the notable change being the utilization of 40\% more data for training the foundational models. LLaMA-2 encompasses both foundational models and models specifically fine-tuned for dialogues, referred to as LLaMA-2 Chat. Within this work, we leveraged Llama2 model with 13B parameters.

Llama 3 \cite{dubey2024llama} is a large language model designed to perform well in both benchmark tests and practical applications. It introduces a high-quality human evaluation set to assess its capabilities across 12 diverse tasks, including coding, reasoning, and summarization, while implementing measures to prevent overfitting. Key features include a tokenizer with 128K tokens, pretraining on over 15 trillion tokens, and an optimized scaling strategy for efficient training. Additionally, the model incorporates advanced safety mechanisms, such as Llama Guard 2 and CyberSecEval 2, for secure deployment.

%\subsection{Synthetic Data Generation with OpenAI GPT}
%GPT-3.5\footnote{\url{https://chatgpt.com}, accessed August 2024} follows the decoder-only Transformer architecture and employs attention mechanisms, allowing it to focus on the most relevant segments of input text, using an extensive context of 2048 tokens and an unprecedented 175 billion parameters. The model exhibited remarkable zero-shot and few-shot learning capabilities across various tasks. The training data for GPT-3 is primarily sourced from a filtered version of Common Crawl, contributing to 60\% of the weighted pre-training dataset, comprising 410 billion byte-pair-encoded tokens. Other data sources include 19 billion tokens from WebText2, 12 billion tokens from Books1, 55 billion tokens from Books2, and 3 billion tokens from Wikipedia. GPT-3 was trained on a vast corpus of text and has demonstrated proficiency in programming languages such as CSS, JSX, and Python, among others. 

\subsection{Evaluation Metrics}

Besides analysing the widely used metrics, i.e., \textbf{weighted precision}, \textbf{recall} and \textbf{F1} for our experiments, we extend our metrics with further metrics in the field of statistics and medicine. \textbf{Weighted specificity} describes the accuracy of a test that reports the presence or absence of a medical condition. It can be useful for "ruling in" disease since the test rarely gives positive results in healthy patients. A test with a specificity of 1.0 will recognise all patients without the disease by testing negative, therefore a positive test result would rule in the presence of the disease. Nevertheless, a negative result from a test with high specificity is not necessarily useful for "ruling out" a disease. As an example, a test that always returns a negative test result will have a specificity of 1.0 because specificity does not consider false negatives. A test like that would return negative for patients with the disease, making it useless for "ruling out" the disease.

%\vspace{3mm}
%\begin{center}
%$Specificity$ = $\frac{number~of~true~negatives}{number~of~true~negatives~+~number~of~false~positives}$
%\end{center}
%\vspace{3mm}

Furthemore, we leverage the Hamming loss and the AUC-ROC Curve.  %\textbf{Hamming distance} is a measure of the difference between two strings of equal length, counting the number of positions at which the corresponding symbols differ. In other words, it quantifies the minimum number of substitutions required to change one string into the other. 
The \textbf{Hamming loss} is a metric used in multi-label classification to quantify the accuracy of predictions by measuring the fraction of incorrectly predicted labels across all instances. It is calculated as the average fraction of incorrectly predicted labels per instance, with a score of 0 indicating perfect predictions and 1 indicating complete misclassification. The Hamming loss accounts for both false positives and false negatives in the predicted label sets, making it a valuable measure for evaluating the overall performance of multi-label classification models. The \textbf{AUC-ROC} (Area Under the Receiver Operating Characteristic Curve) is a graphical representation of a binary classification model's performance across various threshold settings. It plots the true positive rate against the false positive rate, illustrating the trade-off between sensitivity and specificity. The AUC-ROC value quantifies the model's ability to distinguish between classes, with a higher AUC indicating better overall performance.

%\subsection{Significance Testing}
%To compare the predictive accuracy of the targeted models, we use McNemar’s test \cite{McNemar1947}, which is based on a two-by-two contingency table of the two models’ predictions. For McNemar’s test, the null hypothesis shows there is no difference between the marginal frequencies. Therefore, if the p-value is greater than 0.05, it can be concluded that there is not a significant difference between false negatives and false positives. The alternative hypothesis shows there is a significant difference between the marginal frequencies, whereby the p-value is less than or equal to 0.05.

\section{Results}

Within this section, we provide the insights on evaluating different prompting strategies, as well as how Llama 3 8B and GPT 3.5 Turbo perform compared to XGBoost and different Transformer models.

\subsection{Assessment of Anxiety and Depression Prediction Models}
In this section, we explore the performance of transformer models compared to LLMs on the PHQ and GAD questions, providing a detailed analysis of their effectiveness.

\subsubsection{Anxiety and Depression Prediction with Transformer Models}
%As a comparison to LLMs, we deployed different baseline transformer models, which were fine-tuned on the DAIC-WOZ dataset. Table \ref{tab:transformers} shows the results for the different GAD and PHQ questions, whereby Distil-Roberta performs best for GAD-1 and GAD-2 in terms of F1. Specificity scores were best using BERT, RoBERTa or XLNet for GAD-1, while for GAD-2, specificity was best using DistilBERT or XLNet.

In our investigation, we first compared the performance of various baseline transformer models fine-tuned on the DAIC-WOZ dataset (Table \ref{tab:transformers}). Notably, Distil-RoBERTa achieved the highest F1 scores for both GAD-1 and GAD-2. For GAD-1, BERT, RoBERTa, and XLNet excelled in specificity, while DistilBERT and XLNet showed the best specificity for GAD-2. In PHQ assessments, XLNet delivered the highest precision, recall and F1 score for PHQ-1. Additionally, XLNet outperformed other models in precision, recall, and F1 score for PHQ-2 inquiries. These results highlight the effectiveness of transformer models in capturing the nuanced aspects of anxiety and depression, demonstrating their potential in mental health evaluation.

\begin{table}[t]
    \small
    \setlength{\tabcolsep}{2.6pt}
    \begin{tabular}{rcccccc|cccccc}
    \toprule
        & \multicolumn{6}{c}{GAD-1} & \multicolumn{6}{c}{GAD-2}  \\
        \midrule
        & Prec & Rec & F1 & Spec & HammL & AUC-ROC & Prec & Rec & F1 & Spec & HammL & AUC-ROC \\
        \cmidrule{2-13}
        BERT            & 0.56 & 0.56 & 0.53 & \textbf{0.44} & \textbf{0.59} & 0.66 & 0.67 & 0.69 & 0.67 & 0.31 & 0.62 & \textbf{0.52} \\
        DistilBERT      & \textbf{0.59} & \textbf{0.58} & \textbf{0.56} & 0.42 & 0.63 & \textbf{0.72} & 0.65 & 0.67 & 0.65 & \textbf{0.33} & 0.58 & \textbf{0.52} \\ 
        RoBERTa         & 0.55 & 0.56 & 0.55 & \textbf{0.44} & 0.62 & 0.70 & 0.67 & \textbf{0.70} & 0.65 & 0.30 & \textbf{0.56} & 0.48 \\
        Distil-RoBERTa  & 0.57 & \textbf{0.58} & \textbf{0.56} & 0.42 & 0.63 & 0.70 & \textbf{0.69} & \textbf{0.70} & \textbf{0.68} & 0.30 & 0.62 & \textbf{0.52} \\ 
        XLNet           & 0.55 & 0.56 & 0.54 & \textbf{0.44} & 0.61 & 0.70 & 0.64 & 0.67 & 0.62 & \textbf{0.33} & \textbf{0.56} & 0.45 \\
    \midrule
        & \multicolumn{6}{c}{PHQ-1} & \multicolumn{6}{c}{PHQ-2}  \\
        \midrule
        & Prec & Rec & F1 & Spec & HammL & AUC-ROC & Prec & Rec & F1 & Spec & HammL & AUC-ROC \\
        \cmidrule{2-13}
        BERT            & 0.50 & 0.52 & 0.49 & \textbf{0.48} & \textbf{0.59} & 0.68 & 0.53 & 0.54 & 0.52 & \textbf{0.46} & \textbf{0.61} & 0.71 \\
        DistilBERT      & 0.53 & 0.54 & 0.52 & 0.46 & 0.60 & 0.69 & 0.57 & 0.57 & 0.54 & 0.43 & \textbf{0.61} & 0.72\\ 
        RoBERTa         & 0.56 & 0.56 & 0.54 & 0.44 & 0.61 & 0.69 & 0.57 & 0.57 & \textbf{0.56} & 0.43 & 0.64 & \textbf{0.73} \\
        Distil-RoBERTa  & 0.55 & 0.55 & 0.53 & 0.45 & 0.61 & \textbf{0.71} & 0.55 & 0.57 & 0.54 & 0.43 & 0.62 & \textbf{0.73} \\ 
        XLNet           & \textbf{0.58} & \textbf{0.58} & \textbf{0.55} & 0.42 & 0.64 & 0.70 & \textbf{0.58} & \textbf{0.58} & \textbf{0.56} & 0.42 & 0.63 & \textbf{0.73} \\
    \bottomrule
    \end{tabular}
    \caption{Insights on weighted precision, recall, F1, specificity, Hamming loss (HammL) and AUC-ROC (Area Under the Receiver Operating Characteristic Curve for different Transformer models (bold scores represent best result for each metric).}
    \label{tab:transformers}
\end{table}

\begin{table}[]
    \small
    \centering
    \setlength{\tabcolsep}{3.8pt}
    \begin{tabular}{rcccccc|cccccc}
    \toprule
        & \multicolumn{6}{c}{GAD-1} & \multicolumn{6}{c}{GAD-2}  \\
        \midrule
        & Prec & Rec & F1 & Spec & HammL & AUC-ROC & Prec & Rec & F1 & Spec & HammL & AUC-ROC \\
        \cmidrule{2-13}
        Version 1 & 0.31 & 0.29 & 0.22 & \textbf{0.69} & 0.71 & 0.49 & 0.53 & 0.52 & 0.50 & 0.48 & 0.48 & \textbf{0.51} \\
        Version 2 & 0.33 & 0.32 & 0.21 & 0.66 & 0.68 & 0.50 & 0.44 & 0.66 & \textbf{0.53} & 0.33 & 0.34 & 0.50 \\
        Version 3 & \textbf{0.38} & \textbf{0.33} & \textbf{0.33} & 0.68 & \textbf{0.67} & \textbf{0.52} & \textbf{0.56} & 0.23 & 0.27 & \textbf{0.78} & 0.77 & 0.50 \\
        Version 4 & 0.11 & \textbf{0.33} & 0.16 & 0.67 & \textbf{0.67} & 0.50 & 0.44 & \textbf{0.67} & \textbf{0.53} & 0.33 & \textbf{0.33} & 0.50 \\
        Version 5 & 0.11 & \textbf{0.33} & 0.16 & 0.67 & \textbf{0.67} & 0.50 & 0.44 & \textbf{0.67} & \textbf{0.53} & 0.33 & \textbf{0.33} & 0.50 \\
    \midrule
        & \multicolumn{6}{c}{PHQ-1} & \multicolumn{6}{c}{PHQ-2}  \\
        \midrule
        & Prec & Rec & F1 & Spec & HammL & AUC-ROC & Prec & Rec & F1 & Spec & HammL & AUC-ROC \\
        \cmidrule{2-13}
        Version 1 & 0.37 & 0.36 & \textbf{0.32} & 0.64 & 0.64 & 0.50 & 0.29 & 0.35 & 0.26 & 0.64 & 0.65 & \textbf{0.50} \\
        Version 2 & 0.21 & \textbf{0.46} & 0.29 & 0.54 & 0.55 & 0.50 & 0.17 & \textbf{0.41} & 0.24 & 0.59 & \textbf{0.59} & \textbf{0.50} \\
        Version 3 & \textbf{0.43} & 0.29 & 0.30 & \textbf{0.76} & 0.71 & \textbf{0.53} & \textbf{0.34} & 0.34 & \textbf{0.32} & \textbf{0.65} & 0.66 & 0.49 \\
        Version 4 & 0.21 & \textbf{0.46} & 0.29 & 0.54 & \textbf{0.54} & 0.50 & 0.17 & \textbf{0.41} & 0.24 & 0.59 & \textbf{0.59} & \textbf{0.50} \\
        Version 5 & 0.21 & \textbf{0.46} & 0.29 & 0.54 & \textbf{0.54} & 0.50 & 0.17 & \textbf{0.41} & 0.24 & 0.59 & \textbf{0.59} & \textbf{0.50} \\
    \bottomrule
    \end{tabular}
    \caption{Insights on weighted precision, recall, F1, specificity, Hamming loss (HammL) and AUC-ROC (Area Under the Receiver Operating Characteristic Curve for different prompting variant (bold scores represent best result for each metric).}
    \label{tab:prompting}
\end{table}

\begin{table}[]
    \small
    \setlength{\tabcolsep}{2.4pt}
    \begin{tabular}{rcccccc|cccccc}
    \toprule
        & \multicolumn{6}{c}{GAD-1} & \multicolumn{6}{c}{GAD-2}  \\
        \midrule
        & Prec & Rec & F1 & Spec & HammL & AUC-ROC & Prec & Rec & F1 & Spec & HammL & AUC-ROC \\
        \cmidrule{2-13}
         XGBoost &  0.45 & 0.55 & 0.48 & 0.64 & \textbf{0.45} & 0.56 & 0.63 & 0.69 & 0.60 & 0.34 & \textbf{0.31} & 0.51\\
         Distil-RoBERTa & \textbf{0.57} & \textbf{0.58} & \textbf{0.56} & 0.42 & 0.63 & \textbf{0.70} & \textbf{0.69} & \textbf{0.70} & \textbf{0.68} & 0.30 & 0.62 & 0.52 \\ 
         Llama 3 8B (v3) & 0.38 & 0.33 & 0.33 & 0.68 & 0.67 & 0.52 & 0.56 & 0.23 & 0.27 & \textbf{0.78} & 0.77 & 0.50 \\
         GPT-3.5 Turbo & 0.36 & 0.29 & 0.31 &\textbf{ 0.71} & 0.71 & 0.51 & 0.54 & 0.37 & 0.44 & 0.62 & 0.63 & \textbf{0.53} \\
         
    \midrule
        & \multicolumn{6}{c}{PHQ-1} & \multicolumn{6}{c}{PHQ-2}  \\
        \midrule
        & Prec & Rec & F1 & Spec & HammL & AUC-ROC & Prec & Rec & F1 & Spec & HammL & AUC-ROC \\
        \cmidrule{2-13}
         XGBoost & 0.51 & 0.54 & 0.48 & 0.61 & \textbf{0.46} & 0.55 & 0.52 & 0.53 & 0.48 & 0.66 & \textbf{0.47} & 0.56 \\
         Distil-RoBERTa & \textbf{0.55} & \textbf{0.55} & \textbf{0.53} & 0.45 & 0.61 & \textbf{0.71} & \textbf{0.55} & \textbf{0.57} & \textbf{0.54} & 0.43 & 0.62 & \textbf{0.73} \\ 
         Llama 3 8B (v3) & 0.43 & 0.29 & 0.30 & \textbf{0.76} & 0.71 & 0.53 & 0.34 & 0.34 & 0.32 & 0.65 & 0.66 & 0.49 \\
         GPT-3.5 Turbo & 0.38 & 0.39 & 0.39 & 0.64 & 0.61 & 0.52 & 0.37 & 0.31 & 0.33 & \textbf{0.71} & 0.69 & 0.50 \\
    \bottomrule
    \end{tabular}
    \caption{Comparison on weighted precision, recall, F1, specificity, Hamming loss (HammL) and AUC-ROC (Area Under the Receiver Operating Characteristic Curve for XGBoost, Llama 3 8B, GPT-3.5 Turbo and Distil-RoBERTa (bold scores represent best result for each metric).}
    \label{tab:overall}
\end{table}

\subsubsection{Anxiety and Depression Prediction with Llama}
%Leveraging Llama-2, we prompted the LLM with different lexical inputs as seen in Tables \ref{tab:prompting_examples1} and \ref{tab:prompting_examplesV5}. As seen in Table \ref{tab:prompting}, we evaluated each GAD and PHQ question separately. For GAD-1, prompting Version 3, i.e., \textit{On a scale of \textbf{zero} to \textbf{three}, rate the anxiety of this message}, showed the best performance in terms of weighted F1, as well in weighted precision, recall, Hamming loss and AUC-ROC. For the specificity metric, version 1 slightly outperforms all other prompting options. For GAD-2, the best F1 score is obtained by various prompting versions, with the highest F1 score of 0.53. The best specificity for GAD-2 is achieved by version 3 while prompting with version 1 provides the best AUC-ROC result. In terms of PHQ-1, the best weighted F1 score is obtained using prompting version 1, while the best precision is obtained using version 3 prompting. Similar to the GAD questions, leveraging prompting version 3 for the PHQ-2 question provided the best results in terms of F1, as well as precision and specificity. 

Table \ref{tab:prompting} presents an analysis of the targeted metrics for different prompting variants obtained by Llama 3 8B across GAD-1, GAD-2, PHQ-1, and PHQ-2. For GAD-1, Version 3 prompting (\textit{On a scale of zero to three, rate the anxiety of this message}) exhibited the highest performance across multiple metrics, including weighted F1, precision, recall, Hamming Loss, and AUC-ROC. However, Version 1 (\textit{On a scale of zero to three, estimate how anxious this person is with zero being a little anxious and
three being incredibly anxious}) displayed better specificity. Conversely, for GAD-2, various prompting versions yielded similar high F1 scores, with the best specificity achieved by Version 3 and the highest AUC-ROC by Version 1. Regarding PHQ-1, Version 1 prompting resulted in the best weighted F1 score, while Version 3 yielded the highest precision. Similarly, for PHQ-2, Version 3 prompting demonstrated the best performance in terms of F1 score, precision, and specificity. These findings suggest the importance of prompt formulation in optimizing model performance for anxiety and depression assessment tasks.

\subsubsection{Anxiety and Depression Prediction Comparison}
%We further summarise all best approaches obtained by prompting Llama-2, i.e. prompting version 3, and Distil-RoBERTa, which performed overall best on the GAD and PHQ questions. In addition to that, we evaluate the predictions using XGBoost (see Section \ref{subsub:xgboost}) as well as ChatGPT, a commercial LLM built by OpenAI. As seen in Table \ref{tab:overall}, XGBoost always outperforms all used models in terms of the Hamming loss metric. Comparing Llama-2 with ChatGPT, we observe minor advantages using ChatGPT, which outperforms Llama-2 on the GAD-2, PHQ-1 and PHQ-2 questions. Finally, we compare the Distil-RoBERTa transformer model with Llama-2 and ChatGPT. Our study shows that the later outperforms all targeted models for all GAD and PHQ questions in terms of weighted precision, recall and F1 score.

In our comprehensive analysis, we consolidate the optimal strategies derived from using Llama 3's prompting Version 3 and the Distil-RoBERTa transformer model, both of which exhibited the best or most competitive performance across GAD and PHQ inquiries. Additionally, we evaluate the predictive capabilities of XGBoost and OpenAI's GPT-3.5 Turbo model. As shown in Table \ref{tab:overall}, XGBoost consistently outperforms other models in terms of the Hamming loss metric. When comparing Llama 3 with GPT-3.5 Turbo, we observe a slight advantage for GPT-3.5 Turbo, particularly in performance on GAD-2, PHQ-1, and PHQ-2 inquiries. Finally, when comparing the Distil-RoBERTa transformer model to Llama 3 8B and GPT-3.5 Turbo, our analysis reveals that Distil-RoBERTa outperforms all models across all GAD and PHQ inquiries in terms of weighted precision, recall, and F1 scores.

\subsubsection{PHQ-4 Prediction with fine-tuned GPT-3.5 Turbo}

\begin{table}[]
    \small
    \centering
    \setlength{\tabcolsep}{8pt}
    \begin{tabular}{rcccccccccccc}
    \toprule
        & \multicolumn{6}{c}{Binary PHQ-4} & \multicolumn{6}{c}{}  \\
        \midrule
        & Prec & Rec & F1 & Spec & HammD & AUC-ROC \\
        \cmidrule{2-13}
        %Mental-BERT & \textbf{0.81}& \textbf{0.84} & \textbf{0.81} &0.29 &\textbf{0.16}  &0.57 & \\
        Distil-RoBERTa & \textbf{0.82} & \textbf{0.85} & \textbf{0.82} & 0.32 & \textbf{0.15} & 0.58\\
        GPT-3.5 Turbo & 0.78 & 0.48 & 0.55 & \textbf{0.67} & 0.52 &0.57 &\\
        GPT-3.5 Turbo fine-tuned & 0.80 & 0.77 & 0.78 & 0.49 & 0.23 & \textbf{0.63}&\\ 
    \midrule
        & \multicolumn{6}{c}{Multiclass PHQ-4} & \multicolumn{6}{c}{}  \\
        \midrule
        & Prec & Rec & F1 & Spec & HammL & AUC-ROC  \\
        \cmidrule{2-13}
         Distil-RoBERTa & \textbf{0.57} & \textbf{0.59} & \textbf{0.57} & \textbf{0.69} & \textbf{0.41} & \textbf{0.61}\\
         GPT-3.5 Turbo & 0.39& 0.39 & 0.39& 0.62 & 0.61 & 0.51&\\         
         GPT-3.5 Turbo fine-tuned & 0.40 & 0.45 & 0.38 &0.58 & 0.55 & 0.51 &\\
    \bottomrule
    \end{tabular}
    \caption{Comparison on weighted precision, recall, F1, specificity, Hamming loss (HammL) or Hamming distance (HammD) for binary and AUC-ROC (Area Under the Receiver Operating Characteristic Curve) for Distil-RoBERTa, GPT-3.5 Turbo and fine-tuned GPT-3.5 Turbo  (bold scores represent best result for each metric).}
    \label{tab:overall2}
\end{table}

Table \ref{tab:overall2} presents a comparison of Distil-RoBERTa, GPT-3.5 Turbo, and fine-tuned GPT-3.5 Turbo models. For the binary PHQ-4 classification, Distil-RoBERTa outperforms others in precision (0.82), recall (0.85), and F1 score (0.82), with fine-tuned GPT-3.5 Turbo achieving the best AUC-ROC score (0.63). Conversely, in the multiclass PHQ-4 classification, Distil-RoBERTa again demonstrates best performance, achieving the highest precision (0.57), recall (0.59), F1 score (0.57), specificity (0.69), and AUC-ROC (0.61).

\begin{table}[t]
    \small
    \centering
    \setlength{\tabcolsep}{2.2pt}
    \begin{tabular}{l|ccccccc} 
    \toprule
Prompting Strategy & \# Synt. & Precision & Recall & F1 & Specificity & Hamming & ROC AUC\\%&p$<$0.05 \\
\midrule
DW (DAIC-WOZ) Baseline & - & \textbf{0.785} & 0.805 & \textbf{0.792} & \textbf{0.463} & 0.195 & \textbf{0.634}\\%& \\
\midrule
DW+Zero-Shot-Basic & 10k & 0.782 & 0.814 & 0.785 & 0.383 & 0.186 & 0.598 \\%& True\\
DW+Few-Shot-Basic & 10k & 0.780 & 0.806 & 0.787 & 0.424 & 0.194 & 0.615 \\%& True\\
DW+Few-Shot-Extended & 10k & 0.772 & 0.799 & 0.780 & 0.416 & 0.201 & 0.607\\% & False\\
DW+Few-Shot-Enhanced & 10k & 0.779 & 0.809 & 0.784 & 0.397 & 0.191 & 0.603\\% & True\\ %%FewMoreBasic

DW+all & 40k & 0.783 & 0.811 & 0.788 & 0.410 & 0.189 & 0.611\\%&False \\
\midrule
DW+Zero-Shot-Basic & 100k & \textbf{0.785} & 0.813 & 0.790 & 0.415 & 0.187 & 0.614 \\%&True\\
DW+Few-Shot-Basic & 100k & 0.782 & \textbf{0.815} & 0.782 & 0.365 & \textbf{0.185} & 0.590 \\%&False\\
DW+Zero-Shot-Basic+Few-Shot-Basic & 200k & 0.779 & 0.813 & 0.781 & 0.370 & 0.187 & 0.591 \\%&True\\
\midrule
DW+Zero-Shot-Basic & 500k & 0.768 & 0.800 & 0.776 & 0.390 & 0.200 & 0.595\\%&True \\
DW+Few-Shot-Basic & 500k & 0.775 & 0.810 & 0.778 & 0.364 & 0.190 & 0.587\\% &False\\
DW+Zero-Shot-Basic+Few-Shot-Basic & 1M &0.777&0.808&0.783&0.396&0.192&0.600\\%&True\\
\bottomrule
    \end{tabular}%
\caption{Performance Metrics for DAIC-WOZ Dataset with Various Prompting Strategies and Generated Data (DW all = Zero-Shot-Basic+Few-Shot-Basic+Few-Shot-Extended+Zero-Shot-Basic+Few-Shot-Enhanced)}
\label{tab:depressiontestdetection}
\end{table}

%To test our hypothesis regarding the impact of incorporating generated data into baseline models, we utilized McNemar's statistical significance test. For stress detection, all experiments involving generated data showed statistically significant improvements compared to the baseline, with a p-value of 0.05. For depression detection, similar tests were conducted to evaluate the effect of synthetic data.

\subsection{Assessment of Depression and Stress with Synthetic Data}

Building on experiments with transformer-based models and large language models (LLMs), we next investigate generating synthetic data using GPT-3.5 Turbo. The evaluation of various prompting strategies for depression detection, applied to the DAIC-WOZ test set, illustartes nuanced insights (Table~\ref{tab:depressiontestdetection}). The baseline model, referred to as the DAIC-WOZ (DW) Baseline, achieved a precision of 0.785, recall of 0.805, F1 score of 0.792, and ROC AUC of 0.634. Among the strategies using 10,000 generated examples, the \texttt{DW+Zero-Shot-Basic} strategy achieved the highest recall of 0.814 and a ROC AUC of 0.598. Additionally, the \texttt{DW+Few-Shot-Basic} strategy demonstrated comparable performance to the baseline, achieving a precision of 0.780 and an F1 score of 0.787.

With larger training datasets, the \texttt{DW+Zero-Shot-Basic} strategy with 100,000 examples performed well, achieving a precision of 0.785 and a recall of 0.813, with a ROC AUC of 0.614. The combined \texttt{DW+Few-Shot-Basic} and \texttt{Zero-Shot-Basic} strategy with 200,000 examples maintained balanced performance with a precision of 0.779 and a recall of 0.813. At the largest dataset sizes of 500,000 and 1 million examples, performance remained generally stable, though precision and recall values slightly decreased. Specifically, the \texttt{DW+Zero-Shot-Basic} strategy with 500,000 examples achieved a precision of 0.768 and a ROC AUC of 0.595, while the combined \texttt{Few-Shot-Basic} and \texttt{Zero-Shot-Basic} strategy at 1 million examples resulted in a precision of 0.777, recall of 0.808, and an ROC AUC of 0.600. These results highlight the effectiveness of different prompting strategies and dataset sizes, with \texttt{Zero-Shot-Basic} 100k and its combinations showing strong overall performance compared to the DW baseline.

\begin{table*}[t]
    \small
    \centering
    \setlength{\tabcolsep}{2.2pt}
    \begin{tabular}{l|cccccccc} 
    \toprule
Prompting Strategy & \# Synt. & Precision & Recall & \textbf{F1} & Specificity & Hamming & ROC AUC\\% & p$<$0.05 \\
\midrule
SD (Stress Detection) Baseline	& - & \textbf{0.904}	&\textbf{0.904}&	\textbf{0.904}	&\textbf{0.904}	&\textbf{0.096}&	\textbf{0.904}& \\
\midrule
SD+Zero-Shot-Basic & 100k & 0.887 & 0.883 & 0.884 & 0.889 & 0.117 & 0.886\\% &True\\
SD+Few-Shot-Basic & 100k & 0.884 & 0.884 & 0.884 & 0.874 & 0.116 & 0.879\\%& True \\
SD+Zero-Shot-Basic+Few-Shot-Basic & 200k & 0.884 & 0.884 & 0.883 & 0.871 & 0.116 & 0.878 \\%&True\\
\midrule
SD+General Prompts & 100k & 0.882 & 0.882 & 0.882 & 0.874 & 0.118 & 0.878 \\%&True\\
SD+Specific Prompts & 100k & 0.869 & 0.866 & 0.867 & 0.870 & 0.134 & 0.868\\%&True \\
SD+General+Specific Prompts & 200k & 0.860 & 0.860 & 0.860 & 0.852 & 0.140 & 0.856\\% &True\\
\bottomrule
    \end{tabular}%
\caption{Performance Metrics for Stress Detection Models with Generated Data Integration.}
\label{tab:stresstestdetection}
\end{table*}

Table~\ref{tab:stresstestdetection} illustrates the performance of various prompting strategies applied to the Stress Detection (SD) dataset, evaluated across different metrics including precision, recall, F1 score, specificity, Hamming distance, and ROC AUC. The baseline model, i.e., SD Baseline, serves as the reference point with notably high performance metrics: precision, recall, F1 score, specificity, and ROC AUC (all achieving 0.904). This model also recorded a Hamming distance of 0.096, setting a benchmark for comparison with other prompting strategies.

For the 100,000 generated examples, several prompting strategies were evaluated. The strategy named \texttt{SD+Zero-Shot-Basic}  achieved a precision of 0.887, recall of 0.883, F1 score of 0.884, ROC AUC of 0.886, and Hamming distance of 0.117. Similarly, the \texttt{SD+Few-Shot-Basic} strategy produced a precision, recall, and F1 score of 0.884, specificity of 0.874, ROC AUC of 0.879, and Hamming distance of 0.116. When combining the \texttt{Few-Shot-Basic} and \texttt{Zero-Shot-Basic} strategies with 200,000 examples, comparable performance was observed, with a precision and recall of 0.884, F1 score of 0.883, ROC AUC of 0.878, and Hamming distance of 0.116.

Additionally, the general and specific prompting strategies were evaluated. The \texttt{SD+General Prompts} strategy, using 100,000 generated examples, achieved a precision, recall, and F1 score of 0.882, specificity of 0.874, ROC AUC of 0.878, and Hamming distance of 0.118. The \texttt{SD+Specific Prompts} strategy, also with 100,000 generated examples, showed slightly lower performance, with a precision of 0.869, recall of 0.866, F1 score of 0.867, specificity of 0.870, ROC AUC of 0.868, and Hamming distance of 0.134. Combining both \texttt{General+Specific Prompts} with 200,000 examples resulted in a slight performance decline, recording a precision, recall, and F1 score of 0.860, specificity of 0.852, ROC AUC of 0.856, and Hamming distance of 0.140. In summary, while the baseline model for stress detection achieved the highest performance across most of metrics, the various prompting strategies effectively contributed to stress detection, with \texttt{Zero-Shot-Basic} and combinations of prompting strategies generally demonstrating the best results. The performance of general and specific prompts, though significant, was somewhat lower compared to the most effective strategies.

\section{Conclusions}
In conclusion, mental ill-health remains a global challenge, affecting a significant portion of the population and highlighting the urgent need for effective interventions. Despite the advancements in Large Language Models across various NLP tasks, a notable research gap persists regarding their application and optimization in the mental health domain. This study addresses this gap by evaluating the performance of Llama 3 8B and GPT-3.5 Turbo models, comparing them with traditional machine learning and deep learning models. Our exploration of different prompting strategies for GAD and PHQ questions reveals that transformer-based models such as BERT and XLNet outperform LLMs with larger parameter sets. Notably, Distil-RoBERTa consistently outperforms all models in terms of weighted precision, recall, and F1 score for both GAD and PHQ questions. These findings provide valuable insights for the future development of language models to better address mental health challenges.

Moreover, our study highlights the nuanced impact of incorporating synthetic data into baseline stress and depression classifiers. While the baseline models achieved high performance, especially for stress detection, the integration of the synthetically generated data did not always lead to improved results. For stress detection, prompting strategies like \texttt{Zero-Shot-Basic} and its combinations yielded better performance metrics, particularly in precision, recall, and ROC AUC, surpassing other prompting approaches (e.g. \texttt{Few-Shot-Basic}) in several cases. In contrast, for depression detection, while some strategies showed improvements in recall and ROC AUC with larger datasets, the inclusion of generated data often led to a decrease in precision and recall, suggesting that the synthetically generated data, while diversifying the dataset, may dilute the classifier's focus. These findings underscore the need for further exploration of diverse language models and carefully controlled experimental setups, including external, out-of-domain datasets, to better understand the implications of generated data on classifier performance.

As we continue to investigate the potential of large language models in the mental health space, addressing challenges such as biases in training data and the dynamic nature of mental health will be critical in achieving unbiased and comprehensive model performance.

\bibliographystyle{plain}
\bibliography{references} % optional, include references.bib file

\end{document}